# Some Ontological Principles for Designing Upper Level Lexical Resources

Nicola Guarino
National Research Council
LADSEB-CNR, Padova, Italy
[guarino@ladseb.pd.cnr.it]

## Abstract

The purpose of this paper is to explore some semantic problems related to the use of linguistic ontologies in information systems, and to suggest some organizing principles aimed to solve such problems. The taxonomic structure of current ontologies is unfortunately quite complicated and hard to understand, especially for what concerns the upper levels. I will focus here on the problem of *ISA overloading*, which I believe is the main responsible of these difficulties. To this purpose, I will carefully analyze the ontological nature of the categories used in current upper-level structures, considering the necessity of splitting them according to more subtle distinctions or the opportunity of excluding them because of their limited organizational role.

## Introduction

Currently, a number of efforts in the area of language engineering are aimed to the development of systems of basic semantic categories (often called "upper-level ontologies"), to be used as main organizational *backbones*, suitable to impose a structure on large lexical repositories. Examples of such systems are the Penman Upper Model (Bateman et al., 1990), recently evolved into the Pangloss ontology (Knight & Luk, 1994) and the Revised Upper Model (Bateman, 1995), the Mikrokosmos ontology (Mahesh, 1996), and the Wordnet (Miller, 1995) upper structure.

Besides strictly linguistic applications, large lexical resources are getting a high potential relevance for information systems, as they are being used not only for NL interfaces, but also for requirements analysis and specification (Burg, 1997), database design (Ambrosio, Métais & Meunier, 1997; Van de Riet, Burg & Dehne, 1998), conceptual models validation and enhancement (Burg & Van de Riet, 1997; van der Vos, Gulla & van de Riet, 1997), conceptual schema integration (Mirbel, 1997), information retrieval and extraction (Guarino, 1997).

These applications impose however strict constraints on the semantics of the hierarchical structures appearing in current upper level ontologies, since in this case a node does not denote a lexical entry, but rather a class of entities of the domain. In other words, while apparently similar in many cases, lexical relations may be different from semantic relations.

The purpose of this paper is to explore some semantic problems related to the use of linguistic ontologies in information systems, and to suggest some organizing principles aimed to solve such problems. See (Guarino, 1998) for a general overview of the role of ontologies in information systems, and for the proposal of *ontology-driven information systems*. See also (Uschold & Gruninger, 1996; Fridman Noy & Hafner, 1997) for an overview of current work in ontology design (not only confined to linguistic ontologies).

Unfortunately, the taxonomic structure of current ontologies is often quite complicated and hard to understand, especially for what concerns the upper levels. This seems to be in contrast with the widespread belief that the upper level of these ontologies should be: i) largely independent of particular applications; ii) possibly language-independent, at least within a common culture; and iii) easily understandable by everybody, in order to be extensively reusable.

This paper intends to show how the theoretical tools of so-called Formal Ontology (Guarino, 1995; Smith, 1998) can help to formulate a number of ontological distinctions and design principles able to produce cleaner, more general, more rigorous and at the same time more understandable top-level ontologies.

I will focus here on a single problem, which is the main responsible, in my opinion, of the difficulties of current upper levels: *ISA overloading*. To this purpose, I will carefully analyze the ontological nature of the categories used in current upper-level structures, considering the necessity of splitting them according to more subtle distinctions or the opportunity of excluding them from an upper-level taxonomy because of their limited organizational role.

After a discussion of some examples, I will briefly present the basic theoretical tools of Formal Ontology, which address the classical ontological issues of philosophical logic: the theory of parts (mereology), the theory of wholes (topology, in a very general sense), the theory of identity, and the theory of dependence. I will discuss then a few organizing principles based on these tools, which can be of help in improving the semantic coherence and the reusability of upper level lexical resources.

## The Main Problem: ISA Overloading

All ontologies are centered on a taxonomy, based on a partial ordering relation named in various ways, like ISA, subsumption, hyperonymy/hyponymy. Such a taxonomy is the main *backbone* of the ontology, which can be "fleshed" with the addition of attributes and other relations among nodes (like meronymy or antonymy). As rather usual, we shall generically call ISA the main taxonomic relation (not to be confused with InstanceOf, which links a node to the class it belongs to and is not a partial order). The problem with ISA when considering linguistic ontologies like WordNet is that it is intended as a *lexical* relation between words, which not always reflects an *ontological* relation between classes of entities of the world.

Although this fact is well known, the tendency to confuse the two aspects is quite high, especially when linguistic ontologies are used for non-linguistic applications.

For instance, a common praxis in linguistic ontologies is to rely on multiple inheritance to represent polysemy. This results in an *overloading* of the role of ISA links, which, as we shall see, may cause serious semantic problems.

To eliminate these problems, the solution I propose in this paper is to pay more attention to the ontological implications of taxonomic choices, limiting ISA links to connect nodes sharing similar *identity criteria*. In this way, the taxonomy reflects a basic ontological structure with a clear semantics, while the extra-information currently encoded by ISA links can be represented by means of specialized links and attributes.

I report below some examples of what I consider as ISA overloading:

*Confusion of senses:*
- A window is both an artifact and a place (Mikrokosmos)

*Reduction of sense:*
- A physical object is an amount of matter (Pangloss)[1]
- An association is a group (WordNet)
- A person is both a physical object and a living thing (Pangloss)
- A communicative event is a physical, a mental, and a social event (Mikrokosmos, Penman)

*Overgeneralization:*
- A place is a physical object (Mikrokosmos, WordNet)
- An amount of matter is a physical object (WordNet)

*Suspect type-to-role link:*
- A person is both a living thing and a causal agent (WordNet)
- An apple is both fruit and food (WordNet)

*Confusion of taxonomic roles:*
- Taxonomy of qualities in Pangloss and Penman
- Separable Entity in Pangloss

The five classes of examples above represent different kinds of ontological misconceptions, which are discussed in turn.

**Confusion of senses.**

The example reported shows a case where different senses of a word are collapsed into a single class, inheriting from multiple superclasses. In Wordnet, two different "synsets" are associated to the term *window*: one is subsumed by *opening*, while the other by *panel*. While the choice of using multiple inheritance to handle polysemy may appear as economical at a first sight, it presents obvious ontological inconveniences: is there an entity that is *both* a panel and an opening? A cleaner solution is to represent the two senses as disjoint concepts, expressing the intrinsic links existing between them by means of extra relations. For instance, an opening needs to exist in order for a window to function properly.

---

[1] It is interesting to compare this with WordNet, where an amount of matter is a physical object.

**Reduction of sense.**

In this case, the ISA link points to an *aspect* of the meaning of a given concept that does not fully account for its identity: a physical object is more than *just* an amount of matter, an association is more than a group of people, as well as a person is more than a physical object. In the latter case, the fact that persons are *very special* physical objects is represented in Pangloss by the extra link to *living thing*, but here the problem is analogous to the previous case, with the difference that there is now a clear order in the two senses represented by means of multiple inheritance: the *body* sense is more basic than the *living being* sense, since the living being *depends* and *is constituted by* its body. In the last example (Fig. 4), three different senses are collapsed together, as if a communicative event was just a physical event with some extra properties.

**Overgeneralization.**

In this case, a single category includes other categories very different in nature, weakening therefore its meaning. For instance, WordNet's gloss for *physical object* is:

"a physical (tangible and visible) entity; 'it was full of rackets, balls and other objects'"

This seems to contradict the examples reported: i) a place is not tangible; ii) *amount of matter* is an *uncountable* category, while the statement reported in the gloss refers to a *countable* notion of physical object. Therefore, these choices force a weaker interpretation of *physical object.*

**Suspect type-to-role link.**

There is something disturbing in the examples reported: a person is *necessarily* a living thing, while her/he *plays the role* of causal agent only when involved in certain events. Analogously for the apple, which is necessarily a fruit while it *can* be a food. As we shall see, *person* and *apple* are *types*, while *causal agent* and *food* are *roles*. On the basis of ontological analysis, I shall bring a reason why forbidding these links.

**Confusion of organizational roles.**

Another source of troubles (less serious however than the previous ones, since it only affects the overall organization) is the tendency to express all the unary properties of a certain class of entities in terms of superclasses to inherit from. Take for instance the hierarchy of qualities in the Penman Upper Level (inherited by Pangloss): all the various formal properties of qualities (like staticity, polarity, etc.) correspond to nodes of the ontology, which of course gets quite tangled. The problem is that there is no distinction between the nodes carrying a major *organizational role* in the taxonomy, and those that simply express a particular property. Another example is the node *decomposable object* in Penman and Pangloss, which subsumes a number of different classes generating a high degree of tangleness. My suggestion is to avoid including these nodes in the taxonomy, relying on attributes to express these properties. In other words, not all unary properties need to be expressed by means of ISA links.

Most of the examples above refer to cases of polysemy represented by means of ISA links. In the following, I

will briefly introduce some theoretical tools which can be adopted to better understand the systematic *ontological structure* underlying polysemy, with a goal similar to that of (Pustejovsky, 1995; Pustejovsky, 1998).

## The Methodology: Formal Ontology

In the current practice, the engineering process of ontology building lacks well-established principles and methodologies. Some methodological proposals have been made in recent times (Bateman et al., 1990; Gruber, 1995; Mahesh, 1996; Uschold & Gruninger, 1996), but most of the ontologies currently in use seem to be the result of a mixture of ad-hoc creativity and naive introspection. Recently, however, a different line of research began to emerge, characterized by a highly interdisciplinary perspective. The philosophical field inspiring this trend is that of *formal ontology*, which has been defined as "the systematic, formal, axiomatic development of the logic of all forms and modes of being" (Cocchiarella, 1991).

In practice, formal ontology can be seen as the *theory of distinctions* within:

- the entities of the world to be included in our domain of discourse, or *particulars*
- the properties and relations used to talk about such entities, or *universals* [2]

The study of formal ontological distinctions can be organized around a number of core theories (or, better, theoretical tools), which have always been at the basis of philosophical research. I shall briefly report here the relevant questions addressed by these theories; see (Smith, 1998) for a further introductory account.

### Theory of parts.

The theory of parts is at the basis of any form of ontological analysis. Relevant questions that must be addressed are:

- What counts as a part of a given entity?
- What are the properties of the parthood relation?
- Are there different kinds of part?

An important example of a theory of parts is given by *extensional mereology*. Much work must be addressed however in order to come up to a satisfactory theory of *intensional mereology*, where integrity and identity are taken into account. See (Simons, 1987) for a thorough reference to the problems of mereology.

### Theory of wholes.

A given entity (like a collection of disconnected pieces) can have parts without being considered as a single whole. The theory of wholes (or theory of *integrity*) studies the ways of connecting together different parts to form a whole. Relevant questions that must be addressed are:

- What counts as a whole? What does make it a whole?
- In which sense are the parts of a whole connected? What are the properties of such a connection relation?
- How is the whole isolated from the background? What are its boundaries?
- What role do the parts play with respect to the whole?

Notice that in order to understand the various forms of part-*whole* relation (Winston, Chaffin & Herrmann, 1987; Artale et al., 1996) the general theory of parts must be supplemented with a theory of wholes. Together, the two theories form what may be called *mereotopology* (Varzi, 1996; Varzi, 1998).

### Theory of identity.

The theory of identity builds up on the theory of parthood and the theory of wholes, studying the conditions under which two entities exhibiting different properties can be considered as the same. Relevant questions that must be addressed are:

- How can an entity change while keeping its identity?
- Do entities have any essential properties?
- Under what conditions does an entity loose its identity?
- Does a change of parts affect identity?
- Does a change of topological or morphological properties affect identity?
- Does a change of "point of view" change the identity conditions?

The last question is especially relevant for our discussion. For instance, consider the classical example of a vase of clay. Should we consider the vase and the clay it is made of as two separate individuals, or just as two different points of view about the same individual? The answer may be difficult, but a careful analysis tells us that the two views imply different identity criteria: when the vase looses its identity by crashing to the floor, the clay is still there. This is because the clay has an *extensional* criterion of identity, since it always coincides with the sum of its parts, while the vase requires a particular arrangement of its parts in order to be a vase. Therefore, we are in presence of two different individuals. See (Hirsch, 1982) for an account of the identity problems of ordinary objects, and (Noonan, 1993) for a collection of philosophical papers in this area.

### Theory of dependence.

The theory of dependence studies the various forms of existential dependence involving specific individuals that belong to different classes. We refer here to the notion of existence as "actual existence", not as "logical existence". In this sense, existence can be represented by a specific predicate (like in (Hirst, 1991)) rather than by a logical quantifier. Relevant questions that must be addressed are:

- Does the actual existence of an individual necessarily imply the actual existence of another specific individual? *(Rigid dependence)*
- Does the actual existence of an individual necessarily imply the actual existence of some individual belonging to a specific class? *(Generic dependence)*
- Does the fact that an individual belongs to a particular class necessarily imply the existence of a different individual belonging to another class? *(Class dependence)*

---

[2] Just for simplicity, we assume that universals are not included in the domain of discourse.

An example of rigid dependence may be the relationship between a person and his/her brain, while the relationship between a person and his/her heart is an example of generic dependence (because the heart can be substituted with another heart, and the identity of the person does not change). Finally, an example of class dependence is the relationship existing between the class "Father" and the class "Child".

## The Role of Identity Criteria

After this quick overview of the main tools of formal ontology, let us focus on the notion of *identity criterion*, which plays a fundamental role in our discussion. Briefly, we can say that an identity criterion (IC) for a property P is a binary relation $I_P$ such that (Noonan, 1993)

$$Px \wedge Py \wedge I_P xy \rightarrow x=y$$

If, for a given property P, we are able to define such an $I_P$, then we say that P *carries an* IC for its instances.

We see therefore that an IC determines a *sufficient* condition for identity. In practice, ICs for classes corresponding to natural language words are difficult or impossible to express. Rather, it is relatively easy (and quite useful) to identify some *necessary* conditions for identity, which allow one to:

- *individuate* an entity as an instance of a class C
- *re-identify* an instance of C across time (persistence)
- *count* the instances of C

It is important to take in mind that the decision of ascribing an IC to a certain class is the result of our *conceptualization* of the world, i.e. of our *ontology*. While we can assume that for the very basic things of our world, like particles of matter, the ICs are basically the same for every human being, being grounded on notions of spatial and temporal continuity somehow "hardwired" in our sensorial apparatus, the presence of *emergent properties* like a particular structure or a particular shape may result in a change of ICs depending on a specific bias or point of view.

After these clarifications, let us see how ICs may affect a taxonomic organization. Consider to this purpose the beautiful example reported in (Lowe, 1989): should *ordered-set* be subsumed by *set*? Obviously not, exactly because the two classes have different ICs (two different ordered-sets may correspond to the same ordinary set). For the same reason, *association* or *queue of people* should not be subsumed by *group*, and *person* should not be subsumed by *physical object* (as a physical object, a body has persistence conditions different from a living being). Yet, all these ISA links exist in WordNet.

In conclusion, Lowe proposes the following principle, which we assume as the basic principle to be adopted for well-founded ontologies:

> "No individual can instantiate both of two sorts if they have different criteria of identity associated with them".
>
> (Lowe, 1989)

Notice that Lowe refers here to the technical notion of *sort,* which is not the same as an arbitrary class or unary property: according to (Strawson, 1959), a sort "supplies a principle for distinguishing and counting individual particulars which it collects", i.e., it carries an IC. The notion of sort will be discussed in greater detail below.

## A Minimal Ontology of Particulars

In order to address the problems discussed in the introduction, we need to make some minimal assumptions about the basic ontological distinctions among particulars and universals, in the light of the methodological suggestions presented above. I quickly present below what I consider the "basic backbone" of the ontology of particulars and universals, with the only purpose of making clear the discussion on the design principles, which is the goal of this paper.

```
Particular
    Location
        Space           (a spatial region)
        Time            (a temporal region)
    Object
        Concrete object
            Continuant  (an apple)
            Occurrent   (the fall of an apple)
        Abstract object (Pythagoras' theorem)
```

Figure 1: The basic *backbone* of the ontology of particulars.

A minimal ontology of particulars is presented in Fig. 1. The first distinction is between objects and locations. Without assuming locations as primitive entities distinct from objects, the notion of object becomes too general for our purpose, and the distinctions within objects become difficult to explain (see the previous remarks on overgeneralization).

A *location* is either a region of (absolute) space or an interval of (absolute) time. Objects are assumed to be *concrete* or *abstract* according to their possibility of having a location in time and/or in space.

Within concrete objects, I assume here for granted the classical distinction between *continuants* and *occurrents*. They correspond to what are usually called *objects* and *events*, but we have already used the term "object" in a more general sense, and this terminology (suggested in (Simons, 1987)) is less ambiguous.

*Continuants* have a location in space, but this location can vary with time. They have spatial parts, but they do not have a temporal location, nor temporal parts. They always have other continuants as parts.

*Occurrents* are "generated" by continuants, according to the ways they behave in time. In order for an occurrent to exist, a specific continuant must *take part* to it. If the continuant changes its identity, the occurrent also changes its identity, so that continuants are *rigidly dependent* on continuants. Examples of occurrents are the change of location of a body, but also the permanence of a body in a given location for a given time (a state occurrence). Occurrents always have other occurrents as parts (continuants take parts to occurrents, but are not part of them). They have a unique temporal location, while their exact spatial location is less obvious, and it is however bound to the location of the participating continuants. For a thorough review of the ontology of occurrents, see (Casati & Varzi, 1996).

*Abstract objects* do not have a location in space or in time. They are included here just for completeness, as we shall not mention them. Notice however that I refer here to abstract *particulars*, such as Pythagoras' theorem or (maybe?) the number "1". Most of entities often classified as abstract objects are however *universals* (see below).

## Ontological Levels

Let us see now how the application of Lowe's principle makes it possible to introduce systematic distinctions among ontological categories that seem to explain – at least *prima facie* – some of the phenomena of *systematic polysemy* discussed in (Pustejovsky, 1995; Buitelaar, 1998), and implicit in the examples of ISA overloading considered before.

Take for instance an animal, which can be conceptualized as an intentional agent, as a biological organism or just as a piece of matter. I argue that, since these concepts imply different ICs, they correspond to *disjoint* categories: three distinct individuals, instances of the three concepts above, share a common spatial location. Since the animal depends on the underlying biological organism, as well as the biological organism depends on the underlying amount of matter, there is an intrinsic order within these categories, which belong therefore to different *ontological levels*. A *dependence relation* links higher levels to lower levels: an animal depends on its body, which depends on body parts having a certain functionality, which depend on pieces of matter having specific properties, and so on. Notice that this dependence is of a *generic* kind: a vase depends on *some* amount of clay, but not necessarily always the same specific clay (it can loose parts, it can be repaired with a different piece of clay).

| Atomic | (a minimal grain of matter) |
|---|---|
| Static | (a configuration) |
| Mereological | (an amount of matter) |
| Physical | |
|   Topological | (a piece of matter) |
|   Morphological | (a cubic block) |
| Functional | (an artifact) |
| Biological | (a human body) |
| Intentional | (a person or a robot) |
| Social | (a company) |

Figure 2: Ontological levels correspond to *disjoint sets* of particulars, according to the different ICs adopted to conceptualize them. Examples are reported in parenthesis.

We shall introduce therefore a classification of ontological levels based on different *kinds* of IC (Fig. 2), corresponding to different sets of *individuation* and *persistence* conditions (Fig. 3).

The idea of levels (or *strata*) of reality is loosely inspired to Husserl's work (Poli, 1996). The notion adopted here, however, is more explicitly related to the theory of identity, and makes use of some notions of mereotopology. The presentation below has to be intended as rather preliminary and deliberately non-technical, and it expands and refines some ideas introduced in (Guarino, 1997).

Each of the levels of Fig. 2 corresponds to a class of ICs, as explained in Fig. 3. These classes are assumed to describe *disjoint* sets of entities, in accordance with Lowe's principle. They are *orthogonal* to the distinctions among particulars introduced in the previous section. In our examples, we refer however mostly to continuants.

| |
|---|
| Atomic |
|   Individuation: minimal size |
|   Persistence: spatio-temporal continuity |
| Static |
|   Individuation: mereological sum of atoms |
|   Persistence: same properties |
| Mereological |
|   Individuation: mereological sum of entities |
|   Persistence: same parts |
| Topological |
|   Individuation: self-connection |
|   Persistence: similar topology |
| Morphological |
|   Individuation: proximity |
|   Persistence: similar shape |
| Functional |
|   Individuation: purpose |
|   Persistence: persistence of functionality |
| Biological |
|   Individuation: presence of life |
|   Persistence: persistence of life |
| Intentional |
|   Individuation: intentional behavior |
|   Persistence: persistence of intentional behavior |
| Social |
|   Individuation: inter-agent connections |
|   Persistence: persistence of inter-agent connections |

Figure 3: different conditions (only *necessary* for identity) associated to ontological levels.

At the *atomic* level, we consider entities having minimal spatial or temporal dimensions, according to a certain granularity dependent on our conceptualization. As mentioned before, we assume spatio-temporal continuity as a necessary condition for the identity of these entities.

At the *static level*, all the non-temporal properties of a particular contribute to its identity: if one of these changes, identity is lost. In this level only very peculiar objects are defined, namely *configurations* of atoms in the case of continuants and *situations* (i.e., occurrences of configurations) in the case of occurrents. Such objects may play a crucial role in a formal axiomatization (see for instance (Borgo, Guarino & Masolo, 1997), since they avoid the introduction of modal framework.

At the *mereological level*, the IC is *extensional*: two entities are the same if and only if they have the same parts (*mereological essentialism*). Regions of space, temporal intervals and amounts of matter belong to this level. The entities belonging to the subsequent levels have an *intensional* criterion of identity, in the sense that mereological identity is neither sufficient nor necessary for identity.

The *physical level* corresponds to ICs bound to spatial configuration of matter (i.e., to topo-morphological properties). It can be split into two separate layers.

At the *topological layer*, the IC is bound to topological properties: for instance, topological self-connection can be considered as a necessary property to maintain identity: a *piece* of matter belongs to this layer, while a (possibly disconnected) *amount* of matter belongs to the mereologi-cal level. The two things are *distinct entities*, since a piece of matter can cease to exist (generating new pieces) while the same amount of matter is still there. A doughnut belongs to the same level, but with a more sophisticated IC, since its identity changes when the hole is destroyed while the dough remains self-connected.

At the *morphological layer*, the IC is bound to morphological properties (or, in general, *gestaltic* properties related to spatial proximity), like spatial shapes or temporal patterns. A change of these properties can influence identity. A cube-shaped block is an example of an instance of this level: if its shape changes (above a certain limit) it is not *the same cube* any more, while still being the same piece of matter. Another example is a constellation, whose IC does not require topological self-connection.

The levels above the physical level are related to ICs bound to the way objects interact with the external world. At the *functional level*, the IC is bound to functional and pragmatic properties: identity is destroyed when functionality is destroyed. At the *biological level*, the IC is bound to properties related to life: identity is destroyed when biological activity ceases. At the *intentional level*, the IC is bound to capability of intentional behavior: identity is destroyed when such capability ceases. At the *social level*, the IC is bound to social rules and conventions involving the interaction of intentional objects. Identity is destroyed when some of these rules change.

In conclusion, we can see how the introduction of ontological levels leads to simpler and (I believe) more understandable taxonomies, as shown by Fig. 4. The costs of this choice are: i) a moderate proliferation (by a constant factor corresponding to the number of levels) of the number of entities in the domain; ii) the necessity to take into account different relations besides ISA, such as co-localization and dependence.

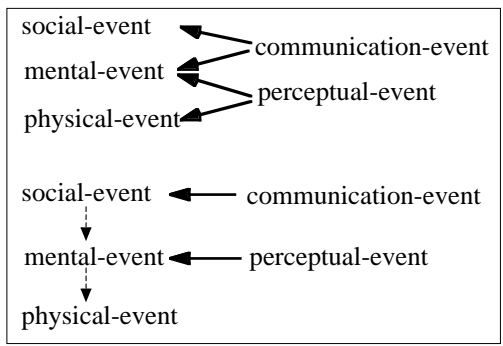

Figure 4: Above: communication and perceptual events in Mikrokosmos. Below: the simplification resulting from the assumption of ontological levels. Dashed arrows denote the dependence relation.

## A Minimal Ontology of Universals

Besides distinctions among particulars, formal ontology addresses also distinctions among *universals*, briefly introduced previously as "the properties and relations used to talk about the entities of our domain of discourse". Before discussing their ontology, we must however clarify that properties and relations are not to be intended here in the usual *extensional* way (i.e., respectively, as subsets of the domain or sets of tuples), but rather in their *intensional* meaning, as conceptual primitives used to model a certain domain. See on this point (Guarino & Giaretta, 1995; Guarino, 1998), where the standard strategy of considering intensional relations as functions from possible world into sets is adopted.

| | | |
|---|---|---|
| Universal | | |
|   Property | | |
|     Type | *(person)* | (+I +R) |
|     Category | *(location, object)* | (-I +R) |
|     Role | | (~R +D) |
|       Material role | *(student)* | (+I) |
|       Formal role | *(patient, part)* | (-I) |
|     Attribution | *(red, decomposable)* | (-I -R -D) |
|   Relation | *(part-of)* | |

Figure 5: The basic *backbone* of the ontology of universals. I = identity, R = rigidity, D = dependence.

A minimal ontology of universals, based on a revision of (Guarino, Carrara & Giaretta, 1994), is reported in Fig. 5. The first distinction is the usual one between *properties* and *relations*, according to the number of arguments. We only focus on *primitive* properties, which are not obtainable by Boolean combination of other properties.

The purpose of studying the distinctions among properties is twofold. On one hand, we are interested in assessing their *organizational role* in a taxonomy, that is their practical relevance as *taxons*, i.e. nodes of a taxonomy; on the other hand, we want to study their attitude to generate clean and understandable hierarchies, with a minimum degree of "tangleness". With the help of formal ontology, we can characterize such distinctions on the basis of the following meta-properties:

1. *Identity* (+I). The property of carrying an IC.
2. *Rigidity* (+R). A property P is rigid if, when P($x$) is true in one possible world, then it is also true in all possible worlds. *Person* and *location* are rigid, while *student* and *tall* are not.
3. *Anti-rigidity* (~R)[3]. A property P is anti-rigid if, for each x, P($x$) is true in one possible world, and false in a different possible world. *Student* and *tall* are both nonrigid (-R) and anti-rigid (~R).
4. *Dependence* (+D). A property P is dependent if, necessarily, whenever P($x$) holds, then Q($y$) holds, with $x \neq y$ and P $\neq$ Q (see the *class dependence* mentioned before). *Father* is dependent, *person* is not.

A *type* is a property that is rigid and carries an IC. Types play the most important organizational role in a taxonomy. Assuming that each type has a distinct set of ICs, we have that, according to Lowe's principle, a taxonomy of types is always a *tree*. When a type specializes another type, it adds further ICs to those carried by the subsuming type. For instance, when the type *triangle* specializes *polygon*, it adds the ICs based on the equivalence of two sides and one angle (or two angles and one side) to those proper of polygons (same sides and same angles).

---

[3] See (Guarino, 1992; Guarino, Carrara & Giaretta, 1994) for a technical account of ontological rigidity and for a characterization of roles as non-rigid entities. The notion of anti-rigidity, introduced here for the first time, seems however to better account for the ontological nature of roles, and explains the issue discussed in Fig. 6.

A *category* is a property that is also rigid but does not carry a specific IC. Since they cannot be subsumed by types (otherwise they would have an IC), categories only appear in the uppermost levels of a taxonomy. Their role is to make clear the most general distinctions.

Types and categories are both rigid, and can be either dependent or independent (*person* is independent, *event* is dependent). A *role* is a property that is *anti-rigid* and is always dependent[4]. *Material* roles like *student* do have an IC, while *formal* roles like *part* do not. However, the IC of material roles is only indirect, since they do not introduce any specific IC, but rather they inherit it from a subsuming type. No explicit mutual disjointness assumption is made for roles, as they tend to generate tangled hierarchies. They have for this reason a limited organizational relevance. It seems therefore advisable to explicitly distinguish roles from types in order to easily isolate the main backbone of a taxonomy, and to perform inferences related to mutual disjointness (Fig. 6). Notice that, as shown by Fig. 6, a role cannot subsume a type, since the former is anti-rigid and the latter is rigid.

Finally, an *attribution*[5] is a property that is not rigid, is not dependent, and does not carry any IC. Attributions do not seem to play any useful organizational role in the ontology of particulars, as they may hold for disparate kinds of entity. Hence, they should not appear as taxons there, while the related information can be contained within the definitions of taxons whose instances exhibit such attribution. We see therefore how with this choice we solve the problems of confusion of organizational roles mentioned at the beginning of this paper.

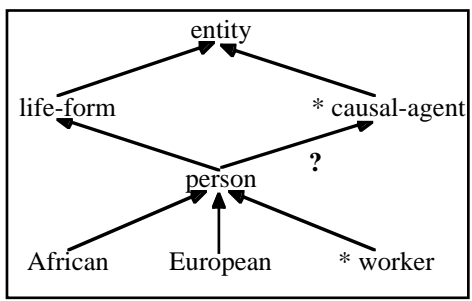

Figure 6. Types and roles in WordNet (roles marked with *). While it is OK for a type to subsume a role, the vice versa is forbidden according to the semantics we have given. Notice that, being types, *African* and *European* are assumed as disjoint.

---

[4] This account of roles in terms of rigidity seems to work well, but it requires some philosophical care with concepts like *child*: in order for *child* to be anti-rigid, there must exist for each person a world where such a person is not a child. This world can be imagined as the one where this person is the first person on the Earth.

[5] This term is introduced in order to avoid confusion with the term *attribute*, largely used in knowledge representation and modelling languages. *Color, part, father* may be attributes, while *red* is an attribution (in this case, an *attribute-value*).

## Conclusions: Some Basic Design Principles

In conclusion, let us summarize the ontology design principles emerging from this discussion, which can solve the *ISA overloading* problems we have mentioned in precedence.

1. *Be clear about the domain*. Any formal theory is a theory about a domain. Such a domain must be clarified in advance. In particular, in our case, it is very important to make clear whether the entities we speak of (i.e., the instances of our classes) are:

- particulars, i.e. either
    - (a) individuals of the actual world;
    - (b) individuals of any possible world (including the actual one);
- universals, i.e. conceptual properties and relations;
- linguistic entities like nouns, verbs or adjectives.

What I have suggested is to have two separate ontologies for particulars (under option (b)) and universals, keeping lexical items out of the domain.

2. *Take identity seriously*. We have seen how the notion of identity criterion (and especially Lowe's principle) can play a crucial role in clarifying ontological distinctions.

3. *Isolate a basic taxonomic structure*. We have seen how the notion of "basic backbone" acquires a rigorous meaning, being constituted by categories and types. Under the assumption of having each one a different set of ICs, types form a tree of mutually disjoint classes. We can reasonably assume, as a design principle, that also categories form a (very shallow) tree of mutually disjoint classes.

4. *Identify roles explicitly*. We have seen that an explicit tag for roles has two advantages: i) you can easily hide them in order to isolate the basic backbone; ii) you can perform inferences involving mutual disjointness while avoiding explicit declarations, unless for cases like *son-daughter*, where two roles are linked by an *antonym* link.

## Acknowledgements

Being this paper yet another (deep) revision of ideas I have been thinking about for many years, I am much indebted to many people. I wish to thank in particular Stefano Borgo, Massimiliano Carrara, Pierdaniele Giaretta, Claudio Masolo, Roberto Poli, Barry Smith, Achille Varzi.